\documentclass[journal]{IEEEtran}

\ifCLASSINFOpdf
\else
\fi

\usepackage{dblfloatfix}

\usepackage{amssymb}
\usepackage{amsmath}
\usepackage{amsthm}

\newtheoremstyle{mystyle}
  {}
  {}
  {\itshape}
  {}
  {\bfseries}
  {.}
  { }
  {}

\theoremstyle{mystyle}

\newtheorem{definition}{Definition}

\usepackage{booktabs}
\usepackage{multirow}
\usepackage{pifont}
\usepackage{tabularx,ragged2e}
\usepackage{makecell}
\usepackage{tikz}
\usepackage{graphicx}
\usepackage{color}
\usepackage{tabu}
\usepackage{cuted}
\usepackage{cite}

\DeclareMathOperator*{\argmax}{arg\,max}

\begin{document}
\title{Destination Prediction by Trajectory Distribution Based Model}

\author{Philippe C. Besse, Brendan Guillouet, Jean-Michel Loubes, and Fran\c{c}ois Royer
\thanks{}}

\markboth{}%
{}

\maketitle
\begin{abstract}
In this paper we propose a new method to predict the final destination of vehicle trips based on their initial partial trajectories. We first review how we obtained clustering of trajectories that describes user behaviour. Then, we explain how we model main traffic flow patterns by a mixture of 2d Gaussian distributions. This yielded a density based clustering of locations, which produces a data driven grid of similar points within each pattern. We present how this model can be used to predict the final destination of a new trajectory based on their first locations using a two step procedure: We first assign the new trajectory to the clusters it mot likely belongs. Secondly, we use characteristics from trajectories inside these clusters to predict the final destination. Finally, we present experimental results of our methods for classification of trajectories and final destination prediction on datasets of timestamped GPS-Location of taxi trips. We test our methods on two different datasets, to assess the capacity of our method to adapt automatically to different subsets. \end{abstract}


\IEEEpeerreviewmaketitle

\section{Introduction}\label{section_introduction}
\IEEEPARstart{M}{onitoring} and predicting road traffic  is  of great importance for traffic managers. With the increase of mobile sensors, such as GPS devices and smartphones, much information is at hand to  understand urban traffic. In the last few years, a large amount of research has been conducted in order to use this data to  model and  analyze road traffic conditions. The aim of this paper is to tackle the issue of predicting the destination of vehicles given a prefix of their trajectory.  This problem has been the subject of a Kaggle challenge entitled "ECML/PKDD 15: Taxi Trajectory Prediction (I)"\cite{kaggleP}. 

The observations are time-stamped locations that correspond to the different positions of vehicles moving within a city monitored at different observation times. When dealing with a dataset composed of trajectories, the difficulty lies in the fact that the data convey both spatial information (locations of the vehicles on the map of the city) and temporal information (for each vehicle, the locations are indexed by time, which creates a sequence of locations that compose a full trajectory). Hence the data have a spatio-temporal structure that must be taken into account in order to model their evolution while the trajectories of the destination points to be predicted are unknown. Vehicle trajectories are also constrained to a road network which makes their time progression very irregular. Locations of vehicles can be seen as two-dimensional data in $\mathbb{R}^2$, that have to be compared to one another, taking into account characteristics of the trajectories they belong to, such as origin and destination.  


In this paper, we propose a method that relies on a distribution based model for the trajectories. We first focus on the temporal structure of the data and gather the locations into clusters of  points that belong to similar trajectories. For this, we rely on a distance that takes into account the geometric properties of trajectories which was developed in  a previous work \cite{Besse2015}. Then, we model  the observations within each obtained cluster by the realisation of a random variable that must be estimated with a distribution on $\mathbb{R}^2$. This estimation step is achieved by considering a mixture of 2-dimensional Gaussian distributions fitted to the data by a maximum likelihood procedure. Thus, each cluster of trajectories corresponds to a parametric distribution model obtained by a mixture of Gaussian distributions. Using this procedure, we obtain a distribution model for points based on the assumption that they belong to a cluster of trajectories. Forecasting the destination of vehicles is a two step procedure : 
\begin{itemize} 
\item we first attribute the observed path of these vehicles to a cluster of trajectories and \item  then extract from the trajectories within the cluster a feature that stands for the final destination point.
\end{itemize} 
 Hence using the learning set, we obtain a density classification method based on  preliminary trajectories clustering. In addition to the forecast properties of this model that will be analysed on the taxis data set, this methodology provides a probabilistic model for spatio-temporal analysis of vehicle flows. It enables the extraction of distribution mobility patterns in an vehicle transportation system that in order to understand urban mobility and flow. 

 The paper is organised into the following sections. Section \ref{section_data_related_work} is devoted to the presentation of the data and the related work. In Section \ref{section_cluster_trajectories} we present how we obtained trajectories clustering by using a proper distance that takes into account the spatial properties of vehicle trajectories. Section \ref{section_method} describes the issue of clustering points into clusters of trajectories with a mixture of Gaussian distributions and how to use these models to predict the final destination of taxi trip. Finally, in Section \ref{section_result}, we present the experimental results and the performance of our models.

\section{Presentation of the forecast problem and related work} \label{section_data_related_work}
Consider vehicles' location data that consist of locations $p(t) \in \mathbb{R}^2$ observed at several observation times $t$ that may differ for each vehicle's path. As an example of such data, through the paper we will test our procedure using two different datasets. The first contains over 11 million taxi-GPS samples of approximately 500 taxis collected over 30 days in the San Francisco, United-States\cite{cabspotting}. The second contains more than 83 millions taxis-GPS data points describing July 2013 to June 2014 for all 442 taxis in circulation in Porto, Portugal. This dataset has been provided for a Kaggle Competition \cite{kaggleP}. This dataset is also composed of metadata associated to the taxi trips, such that client, taxi stand or taxi identification but the dataset in San Francisco only got the taxi identification. Hence, we deliberately do not use these attributes because we want our method to be easily adapted from one dataset to another. We only use locations in $\mathbb{R}^2$ and the associated timestamp. 

Our goal is to be able to determine the final destination point by observing the beginning of a trajectory. This issue is common to various area and data, animal migrations \cite{Gaffney1999}, VideoFrames applied on Robotic \cite{Kruse1997}, Human \cite{Vasquez2004} Vehicle on crossroads \cite{Hu2006}, or GPS data to study behaviour in Urban Commercial Complexes \cite{Kim2011}. But vehicle trajectories are very different objects, they are constrained to a road network and have very irregular time progressions. The study of destination prediction requires comparing the information of previous trajectories with the current location of trajectories in order to identify the destination. Many authors have already discussed this issue.
In the winning solution of the Kaggle-ECML/PKDD discovery challenge on taxi destination prediction, De Brebrisson \textit{et al.} \cite{Brebisson2015} used a  multi-layer perceptrons neural network on features vector composed of  coordinates of beginnings of trajectories, and diverse context information, such as the departure time, the driver id and client information. Their training set has been built to match the trajectories in the test set's competition. If this solution can easily be adapted to other datasets, it implies a new training for some tested trajectories using more location information. Moreover, neural network scores are hard to interpret and can not be used to better understand the characteristics of the dataset. 
Krumm \textit{et al.} \cite{Krumm2006} and Ziebart \textit{et al.} \cite{Ziebart2008} also used external information, in addition to historical trajectories, such as travel time, trajectory length, accident reports, road condition, and driving habits. They incorporate this information into Bayesian inference to compute the probabilities of predicted destinations. Parteson GRAPH \textit{et al.} \cite{Patterson2003} also used a Bayesian method to predict destination but for specific individuals based on their historical transport modes. The main idea of these studies is to use the external information to enhance the quality of the prediction. It then becomes dependent on the presence of this information and is inapplicable without them. 

Monreale \textit{et al} \cite{Monreale2009} built a decision tree, named T-pattern Tree based on extracted movement patterns and predicted the next location of a new trajectory finding the best matching path in the tree. Tiesyte and Jensen \cite{Tiesyte2008} proposed a nearest-neighbour trajectory method that utilised distance measures to identify the historical trajectory most similar to the current partial trajectory.

Finally, most of the work dealing with destination prediction issue uses probabilistic methods based on the location to identify the most probable location after creating the probability model. Among them, the Markov model has been widely applied in predicting destinations. Ashbrook \textit{et al.} \cite{Ashbrook2003} find potential destinations by clustering GPS data, then predicting destinations from these candidates based on Markov models trained to find the next most likely destination based on those recently visited.  Gambs \textit{et al.} \cite{Gambs2012} determined the destination by using the mobility Markov chains from the sequence of the POI (Point of Interest) to create the model. Simmon \textit{et al.} \cite{Simmons2006} also built the probabilistic model through observation of the drivers' habits. All these studies build prediction based on habits of one or a group of specific individuals based on their historical trips. But they require knowing the identity of the driver. Xue \textit{et al.} \cite{Xue2013} proposed a method which decomposes all available patterns into subtrajectories of neighboring locations. The subtrajectories are assembled into synthesized trajectories. Then, they build the Markov model, which quantifies the correlation between adjacent locations. The main drawback of both bayesian inference and the Markov model is in establishing how well they discretize the space. Either they use the true road network, \cite{Patterson2003, Ashbrook2003, Simmons2006}, which requires significant amount of extra work to map the GPS data to the graph network, or they use a grid of square cells \cite{Krumm2006, Ziebart2008, Xue2013} which is a rough representation of the space and produces results dependent on the choice of the discretization of the grid. 

Choi and Hebert \cite{Choi2006} present a Markov model based on segments of the trajectories, where latent segments are obtained by clustering segments of past trajectories. New trajectories are then modelled as a concatenation of segments, which are assumed as noisy realisations of the latent segment. One of the major drawbacks of this method is that it is used for short term prediction (at most 10 seconds ahead). Wiest \textit{el al.} \cite{Wiest2012} proposed a probabilistic trajectory prediction based on two types of mixture models. They also predict the vehicles trajectories only several seconds into the future.

\section{Model to Cluster Trajectories}\label{section_cluster_trajectories}

In this section we describe how we cluster trajectories. We first recall the definition of trajectory used in this paper.

\begin{definition} A trajectory $T^i$ is defined as  \\
$T^i$ : $[(p^i_1,t^i_1),\hdots,(p^i_{n^i},t^i_{n^i})]$, \\ where $p^i_k  \in \mathbb{R}^2,t_k \in \mathbb{R} ~ \forall k \in [1,\hdots,n^i], ~\forall n^i \in \mathbb{N} $ and $n^i$ is the length of the trajectory $T^i$.
\end{definition}

To cope with the sampling issues of trajectories, we first complete, when required, the locations between $p^i_j$ (at time $t^i_j$) and $p^i_{j+1}$ (at time $t^i_{j+1}$) by the piece wise linear representation between each successive location $p^i_j$ and $p^i_{j+1}$ resulting in a line segment $s^i_j$ between these two points. This new representation is called the \textit{piece wise linear} trajectory. In this representation, no assumption is made about time indexing of segment $s^i_j$.

\begin{definition} A piece wise linear trajectory is defined as 
$T^i_{pl}$ : $(s^i_1,\hdots,s^i_{n^i-1})$ , where $s^i_j = [p^i_j,p^i_{j+1}] \in \mathbb{R}^4$ and $n^i_{pl}$ is the length of the piece wise linear trajectory.
\end{definition}
The length of the PL-trajectory $n^i_{pl}$ is the sum of the lengths of all segments that compose it : $n^i_{pl} = \sum_{j \in [1 \hdots n^i-1]} \|p^i_jp^i_{j+1}\|_2$.

\begin{table}[!ht]\label{table_notation}
\caption{Notation}
\centering
{\tabulinesep=0.9mm
 \begin{tabu}{|c|l|}
\hline
$\mathcal{T}$ & The set of trajectories \\ \hline
$T^i$ & The $i^{th}$ trajectory of set $\mathcal{T}$ \\ \hline
$n^i$ & Number of locations in trajectory $T^i$ \\ \hline
$p^i_j$ & The $j^{th}$ location of $T^i$\\ \hline
$t^i_j$ & The time index of location $p^i_j$\\ \hline
$l^i$ & Label of the $i^{\text{th}}$ Trajectory\\ \hline
$\mathcal{T}^m=\{T^i | l^i=m\}$ & Set of trajectories in cluster $m$ \\ \hline 
$\mathcal{P}^m=\{p^i_j | l^i=m\}$ & Set of points in cluster $m$ \\ \hline 
$\mathcal{C}(\mathcal{T}) = \{\mathcal{T}^1,\hdots,\mathcal{T}^{K}\}  $ & Set of cluster of trajectories \\ \hline
$K $ & Number of clusters of trajectories in $\mathcal{C}(\mathcal{T})$ \\ \hline
$l^i_j$ & Label of the location $p^i_j$  \\ \hline
$\mathcal{P}^m_n = \{p^i_j | l^i_j = m , l^i = n \} $ & Set of points in cluster $(m,n)$  \\ \hline
$\mathcal{C}(\mathcal{P}^m) = \{P^m_1,\hdots,P^m_{k^m}\}$ & Set of points cluster \\ \hline
$k^m$ & Number of clusters of points in $\mathcal{C}(\mathcal{P}^m)$ \\ \hline
$\#S$ & Number of elements in the set S \\
\hline
\end{tabu}}
\end{table}

In a previous paper\cite{Besse2015}, we proposed a method to cluster trajectories based on the behaviours of the users.  This clustering is obtained by \textit{hierarchical clustering} with the \textit{ward} linkage criterion based on a distance between trajectories, the \textit{Symetrized Segment-path-Distance}.

\begin{definition} \textit{Symmetrized Segment-Path Distance} \label{definition-symetrized-distance}
$$ \label{equation_distance_modified_OWD}
D_{SSPD}(T^1,T^2)=\frac{D_{SPD}(T^1,T^2)+D_{SPD}(T^2,T^1)}{2}.
$$
where 
$$ 
D_{SPD}(T^1,T^2)=\frac{1}{n_1}\sum_{i_1=1}^{n_1} D_{pt}(p_{i_1}^1,T^2).
$$
and where $D_{pt}$ is the distance from a point to a trajectory as defined below.
\end{definition}

\begin{definition} $Point-to-Segment$ distance.\label{definition-distance-segment-trajectory}
$$ \small \displaystyle
D_{ps}(p^1_{i_1},s^2_{i_2})=\left\{
   \begin{array}{ll}
       \|p^1_{i_1}p^{1proj}_{i_1}\|_2 & \mbox{if } p^{1proj}_{i_1} \in s^2_{i_2}, \\
       \min(\|p^1_{i_1}p^2_{i_2}\|_2,\|p^1_{i_1}p^2_{i_2+1}\|_2) & \mbox{otherwise.}
   \end{array}
\right.$$
\end{definition}
\textit{Where $p^{1proj}_{i_1}$ is the orthogonal projection of $p^{1}_{i_1}$ on the segment $s^2_{i_2}$}.\\

This distance compares trajectories as a whole, regardless of their time indexing or the number of locations that compose them. It enables us to produce clusters of trajectories describing the traffic flow of the trajectory set $\mathcal{T}$. The partition, $\mathcal{C}(\mathcal{T})$ of $\mathcal{T}$ shows the $K$ main paths taken by the users. The number of clusters $K$, is not fixed. It depends on the dataset, or the precision we want to use in order to described it. In Section \ref{section_result}, we discuss the choice of $K$ for both datasets according to the values of the different quality criteria described in the same Section. 

\section{A Probabilistic Model for Trajectory Classification}\label{section_method}

After obtaining clusters of trajectories that discriminate the main patterns of the traffic flow in the city, we aim to predict the final destination of a vehicle for which we only observe the beginning of its path. Hence, We observe a succession of locations in $\mathbb{R}^2$. To assign these points to a cluster of trajectories, we model the clusters by a mixture of 2d Gaussian distributions. Thus, for each cluster, we obtain a Gaussian likelihood estimated using only the data belonging to this cluster. The observed locations will then be assigned to the most likely cluster according to these different likelihoods.

\subsection{Points Partitioning within Clusters of Trajectories}\label{subsection_trajectory_representation}

Recall that we obtain a set of clusters of trajectories, $\mathcal{C}(\mathcal{T})$. For a new trajectory, we want to  be able to assign it to the cluster it most likely belongs to.
For this purpose, we build a Gaussian mixture model for every cluster of trajectories $ \mathcal{T}^m \in \mathcal{C}(\mathcal{T})$ from the set of all the points $ \mathcal{P}^m$ that compose these trajectories.  A Gaussian mixture model assumes that all points from $ \mathcal{P}^m$ are generated from the sum of $k^m$ Gaussian distributions $\phi$ , which are, in our case, 2-variate Gaussian distributions.

\begin{definition}\label{gm}
A Gaussian Mixture Model is a weighted sum of $k^m$ component Gaussian densities as given by the equation,

$$\Phi^m(p)=\Phi(p|\Theta^m) = \displaystyle \sum_{k=1}^{k^m} \omega^m_k \cdot \phi^m_k(p),$$

where $\omega^m_{k}$ is the mixture weight, \textit{i.e.} the prior probability for any point $p$ belonging to the $k^{\text{th}}$ cluster, such that $\sum_{k=1}^{k^m} \omega^m_{k} =1$, and $\phi^m_k(p)$, $i=1,\hdots, k^m$ are the component Gaussian densities.
\end{definition}

Each component density is a Gaussian density on $\mathbb{R}^2$.

\begin{definition}\label{2dmnd}
The density function of a normal distribution, $\phi^m_k$, is defined as,

\begin{eqnarray*}
\phi^m_k(p) & = & \phi(p|\mu^m_k,\Sigma^m_k) \\
& =& \frac{1}{\sqrt{(2\pi)^{2}|\Sigma^m_k|}} e^{\big[-\frac{1}{2}(p-\mu^m_k)^{tr}\cdot(\Sigma^m_k)^{-1}\cdot(p-\mu^m_k)\big]},
\end{eqnarray*}

where $\mu^m_k \in \mathbb{R}^2 $ and $\Sigma^m_k \in \mathbb{R}^{2x2}$ are respectively the location and the covariance matrix of $\phi^m_k$, and $|\Sigma^m_k|$ is the determinant of the matrix $\Sigma^m_k$.

\end{definition}

$\Theta^m=\{\omega^m_1,\mu^m_1,\Sigma^m_1,\hdots,\omega^m_{k^m},\mu^m_{k^m},\Sigma^m_{k^m}\}$ is the list of parameters of the GMM distribution $\Phi^m$.

To evaluate the parameters $\Theta^m$, we use the maximum likelihood estimation. Its aim is to find the parameters which maximise the likelihood function of $\Phi^m$, given the training set $\mathcal{P}^m$. The \textit{GMM} likelihood, can be defined as,

\begin{equation}\label{gmml}
\mathcal{L}(\Theta^m|\mathcal{P}^m) = \prod_{p \in \mathcal{P}^m} \Phi^m(p).
\end{equation}

The maximum likelihood estimators, $\Theta^m_{ML}$, are the parameters which maximise the \textit{GMM} likelihood function.

\begin{equation}\label{mle}
\Theta^m_{ML} = \argmax_{\Theta} \mathcal{L}(\Theta^m|\mathcal{P}^m).
\end{equation}
 
For each \textit{GMM}, the number of components $k^m$ is set to the value which maximise the criterion information $BIC= -2\ln{ L(\Theta^m|\mathcal{P}^m)}+k\ln(|\mathcal{P}^m|)$.

\begin{equation}\label{bic}
k^m = \argmax_k BIC(k).
\end{equation}

The complete set of trajectories is then modelled by the set of $K$ \textit{GMM}'s, one for each set of points, $\mathcal{P}^m$. Each of these sets has been partitioned into $k^m$ groups : $\mathcal{C}(\mathcal{P}^m) = \{\mathcal{P}^m_1, \hdots, \mathcal{P}^m_{k^m} \mathcal{} \}$. Using this modelling procedures, we obtain several cluster of locations, each one corresponding to a mode of the estimated Gaussian mixture distribution. We got a density based clustering of the cloud points, which produces a data driven grid of similar points within a cluster of trajectories.
Now that we have described the space, we want to use the model to predict the final destination of new trajectories. For that, we want to be able to assign the new trajectory to the cluster it most resembles.

\subsection{Classification of Trajectories}\label{subsection_trajectory_classification}
 Here, We present our method for classification of trajectories. For a new trajectory $T^c$, we want to assign it to the cluster of trajectories it most likely belongs. For this purpose we compute the \textit{simple score}, $s^m(T^c)$ for all the \textit{GMM}s $\Phi^m$. The score is the value of the likelihood function of $\Phi^m$ given the points that compose the trajectory $T^c$. It represents how likely the trajectory $T^c$ belongs to the cluster $m$.

\begin{definition}\label{likelihood_score}
The simple score, $s^m(T^c)$, for a trajectory, $T^c$, to be assigned to the cluster $m$ is defined as:
    $$\begin{aligned}
s^m(T^c)&=& \mathcal{L}( \Theta^m_{ML}|T^c)&= P(T^c | \Theta^m_{ML} ) \\
& & & =\displaystyle \prod_{p^c_j \in T^c} \Phi^m(p^c_j|\Theta^m_{ML}) \\
  \end{aligned}$$
\end{definition}

In this way, we can assigned the trajectory to the cluster with the highest \text{affinity score}.

\begin{equation}\label{simple_class_guess}
l^c_{guess}  = \displaystyle \max_{m \in [1\hdots K]} s^m(T^c).
\end{equation}

We highlight the fact that this method enables us to compute a score for the trajectory and for each cluster. The trajectory is not attributed to one cluster and one cluster only. This is relevant because when only a few points of the trajectory are known, we cannot always be totally certain of the final destination. Several destinations are possible. Hence the score computed is a probability that the trajectory belongs to the cluster of trajectories.

\subsection{Complete Model}\label{subsection_extended_model}

We want to test the influence of auxiliary variables on the quality of our classification method, such hour of the day, or day of the week, during which the trip takes place. The likelihood score as defined in Definition \ref{likelihood_score} does not take into account contextual information. However, we can assume that prior knowledge may help to discriminate the trajectories. Indeed, a path may more likely be taken than an other at a given hour of the day or day of the week. We look forward to verifying this hypothesis by including \textit{auxiliary weights}. For this we define a new \textit{complete score} taking into account the following weights. 

\begin{definition}\label{likelihood_complete_score}
The complete score, $s_{c}^m(T^c)$, for a trajectory, $T^c$, to be assigned to the cluster $m$ is defined as:
$$\begin{aligned}
    s^m_{c}(T^c)&= \mathcal{L}_{ap}(\Theta^m_{ML} | T^c, E^c)\\
    &= P(E^c|\Theta^m_{ML})P(T^c | \Theta^m_{ML} ) \\
    &= \alpha(m,h^c,d^c) \displaystyle \prod_{p^c_j \in T^c} \Phi^m(p^c_j|\Theta^m_{ML})\\
  \end{aligned}$$
\end{definition}

The definition of the \textit{complete score}(\ref{likelihood_complete_score})  is generic. $h^c \in [0,\hdots, 23]$ and $w^c \in [1,\hdots, 7]$ are respectively the hour of the day and the day of the week at which the trajectory $T^c$ begins. This information can be interpreted in different manners which will result in a different value for the \textit{auxiliary weight} $\alpha(m,h^c,w^c)$ according to the information we are taking into account. We define three different weights:
\vspace{2pt}

\begin{itemize}
\setlength\itemsep{10pt}
\item The \textit{Empiric weight} describes the distribution information of the trajectory cluster. $$a_{emp}(m)=\frac{\#\mathcal{T}^m}{\#\mathcal{T}}.$$
\item The \textit{Weekday weight} describes the distribution information of the trajectory cluster at a given day of the week,  $$a_{wd}(d,m)=\frac{\#\{T^c~|~T^c \in \mathcal{T}^m, d^c=d\}}{\#\{T^c~|~T^c \in \mathcal{T}, d^c=d\}}.$$
\item The \textit{Hours weight} describes the distribution information of the trajectory cluster at a given hour of the day,   $$a_{h}(h,m)=\frac{\#\{T^c~|~T^c \in \mathcal{T}^m, h^c=h\}}{\#\{T^c~|~T^c \in \mathcal{T}, h^c=h\}}.$$
\end{itemize}

The \textit{auxiliary weight} $\alpha(m,h^c,w^c)$ is the product of any combination of these weights.

\subsection{Model for Final Destination Prediction}
We present here how our model can be used to predict the final destination of the user trips. We have defined, section \ref{subsection_trajectory_classification}, a \textit{simple score} and a \textit{complete score} for each trajectory to belong to a cluster of trajectory. Hence we can assign the new trajetory to the clusters it most likely belongs. We can then use the information from the trajectories that compose these clusters to predict the final destination of the new trajectory. From this, we define two different methods for predicting final destination.

On the one hand, we consider only the trajectories from the cluster $m$ with the highest score.

 \begin{equation}\label{dpred_1}
 \begin{aligned}
d_{pred_1}(T^c) &= \frac{1}{\#\mathcal{T}^m}\displaystyle \sum_{\substack{i~s.t.\\l^i=m}} p^i_{n^i} \\
        &= d^m,~s.t.~m=l_{guess}^c, \\
  \end{aligned}
\end{equation}
where $d^m$ is the mean of the locations of all final destinations of the trajectories in cluster $\mathcal{T}^m$.

On the other hand, in order to take advantage of the fact that the trajectory $T^c$ is not strictly assigned to one and only one cluster, we define, $d_{pred_2}$, as a weighted sum of the mean final destination of every cluster.

 \begin{equation}\label{dpred_2}
\begin{aligned}
d_{pred}(T^c) &= \sum_{m=1}^K~\frac{s^m(T^c)}{\sum_{k=1}^K s^k(T^c)} \cdot  d^m   \\
              &= \sum_{m=1}^K~s_w^m(T^c) \cdot d^m,   \\
  \end{aligned}
\end{equation}

where $s_w^m(T^c)$ is the \textit{weighted affinity score} of $s^m(T^c)$
The \textit{complete score} can be used instead of the \textit{simple score} to take into account the effect of the auxiliary variable in the final destination prediction.

\section{Experimental Results}\label{section_result}

In this section we present experimental results to evaluate both classification and final destination's prediction methods. To evaluate prediction error, we use the \textit{Haversine Distance} (see Definition \ref{def_haversine}), which is the evaluation metric used in the Kaggle competition \cite{kaggleP}. The Haversine Distance measures distances between two points on a sphere based on their latitude and longitude. We use a 10-cross validation method to calculate this error by learning on $90\%$ the data: the training set $\mathcal{T}_{train}$,  and forecasting the remaining $10\%$: the test set $\mathcal{T}_{test}$. The error forecast is the average for all the training sets. We repeat this operation ten times, such that every set has been considered as the test set, to ensure a more accurate estimation of model prediction performance.

To evaluate our method during trajectory completion, we introduce the definition of a partial trajectory, below. 
A $p$-trajectory, $T^i(p)$ of a trajectory $T^i$ is a subset of this trajectory such that the length of the piecewise representation of $T^i(p)$ is at most p times the size of the length of the piecewise representation of $T^i$.

\begin{definition} The $p$-trajectory $T^i(p),\forall p\in[0,\hdots1]$ is defined as the trajectory: \\ $T^i(p) = ((p^i_1,t^i_1),\hdots,(p^i_{n^i(p)},t^i_{n^i(p)}))$ 
s.t. $\frac{n_{pl}^i(p)}{n^i_{pl}} \leq p$ \\
where $n^i(p)$ is the number of locations that compose the p-trajectory $T^i(p)$.
\end{definition}

\subsection{Data And Clustering Results}

\begin{figure}
\centering
\includegraphics[width=\linewidth]{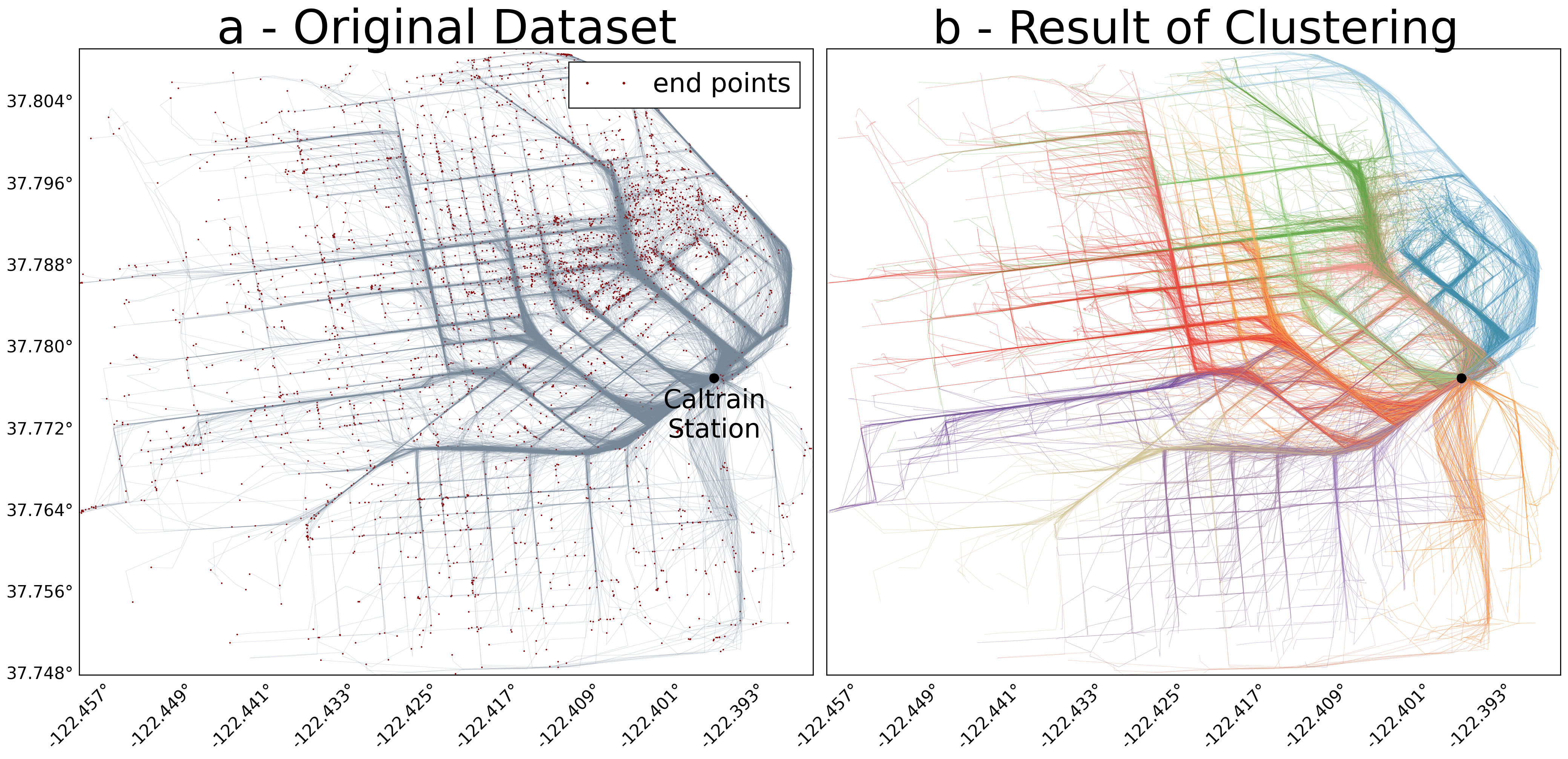}
\caption{Caltrain Station, San Francisco Dataset and its Partitioning in 25 clusters}
\label{sanfrancisco_clustering}
\end{figure}

\begin{figure}
\centering
\includegraphics[width=\linewidth]{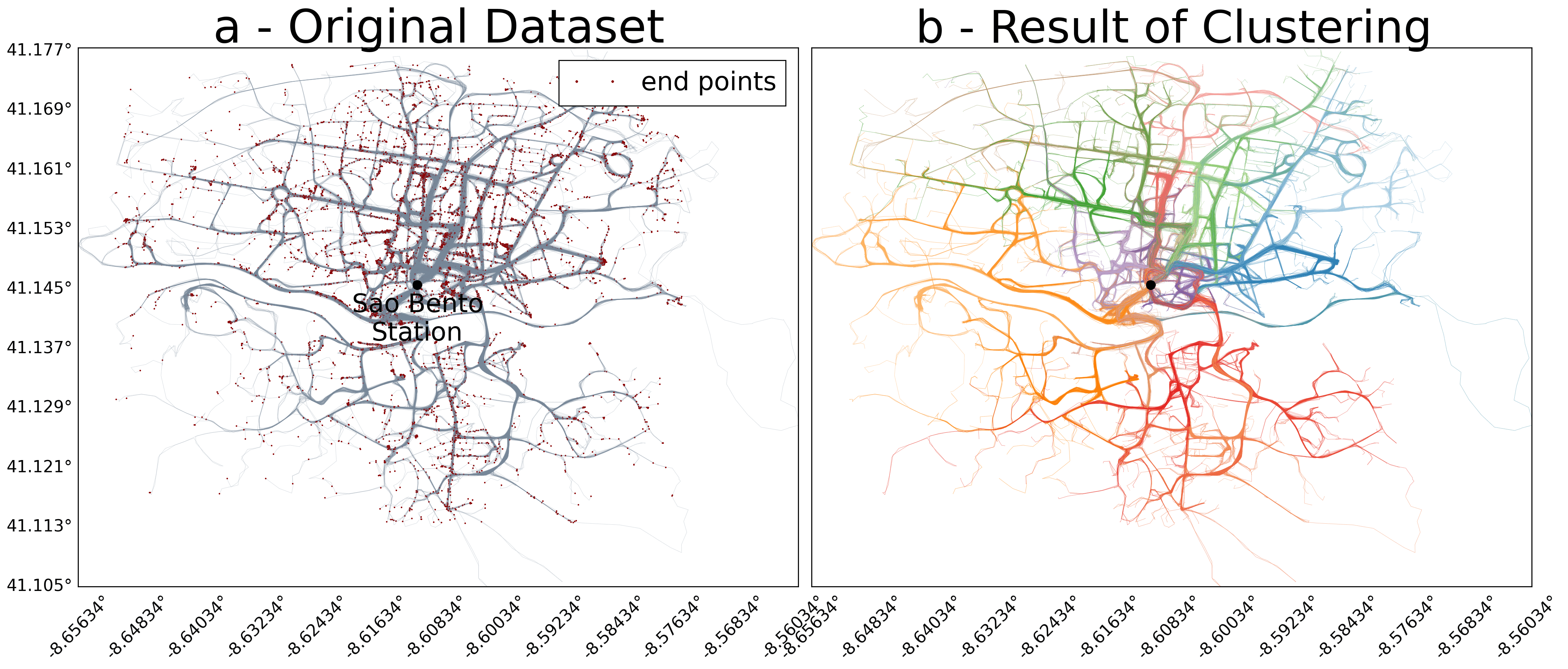}
\caption{Sao Bento Station, Porto Dataset and its Partitioning in 45 clusters}
\label{porto_clustering}
\end{figure}

To analyse our result and test its scalability, we test our model on two different subsets. The first one is a subset of taxi trajectories from San Francisco\cite{cabspotting}. It is composed of $4.127$ trajectories, all starting from the Caltrain Station and ending in an area of size $6,327 \times 6,827$ km in the center of the city.
The second is composed of $19.423$ trajectories from taxis in the center of Porto \cite{kaggleP}, leaving from the Sao Bento Station and ending in a delimited area of size $8,116 \times 8,068$ km.
These two datasets are displayed on the left in Figure \ref{sanfrancisco_clustering} and Figure \ref{porto_clustering}.

These two datasets are quite different. In San-Francisco, the road network looks like a grid, most of the streets are either parallel or perpendicular to each other. In Porto the network is more irregular. These differences will enable us to test the scalability of our method and its capacity to be adapted on different datasets.

On figure \ref{sanfrancisco_clustering} and \ref{porto_clustering} we can see the results of the clustering described in section \ref{section_cluster_trajectories} on the data sets from San Francisco and Porto. We can see that the clustering obtained restuls in a group of trajectories tracing the same path from the selected departure points.

\subsection{Trajectory Classification}\label{subsec_tc}

To evaluate the quality of our classification, we observe the percentage of trajectories that have been assigned to their true cluster.

\begin{definition} The quality criterion, $Q_{class}$, for the classification, is the  percentage of well classified p-trajectories $,\forall p\in[0,\hdots1]$, defined as:
 $$Q_{class}(p) = \frac{\#\{T^i(p) | l^i_{guess}=l^i, T^i \in \mathcal{T}_{test} \}}{\#\{T^i(p) | T^i \in \mathcal{T}_{test} \}}$$
 \end{definition}
 
\begin{figure}
\centering
\begin{minipage}{0.49\linewidth}
\includegraphics[width=\textwidth]{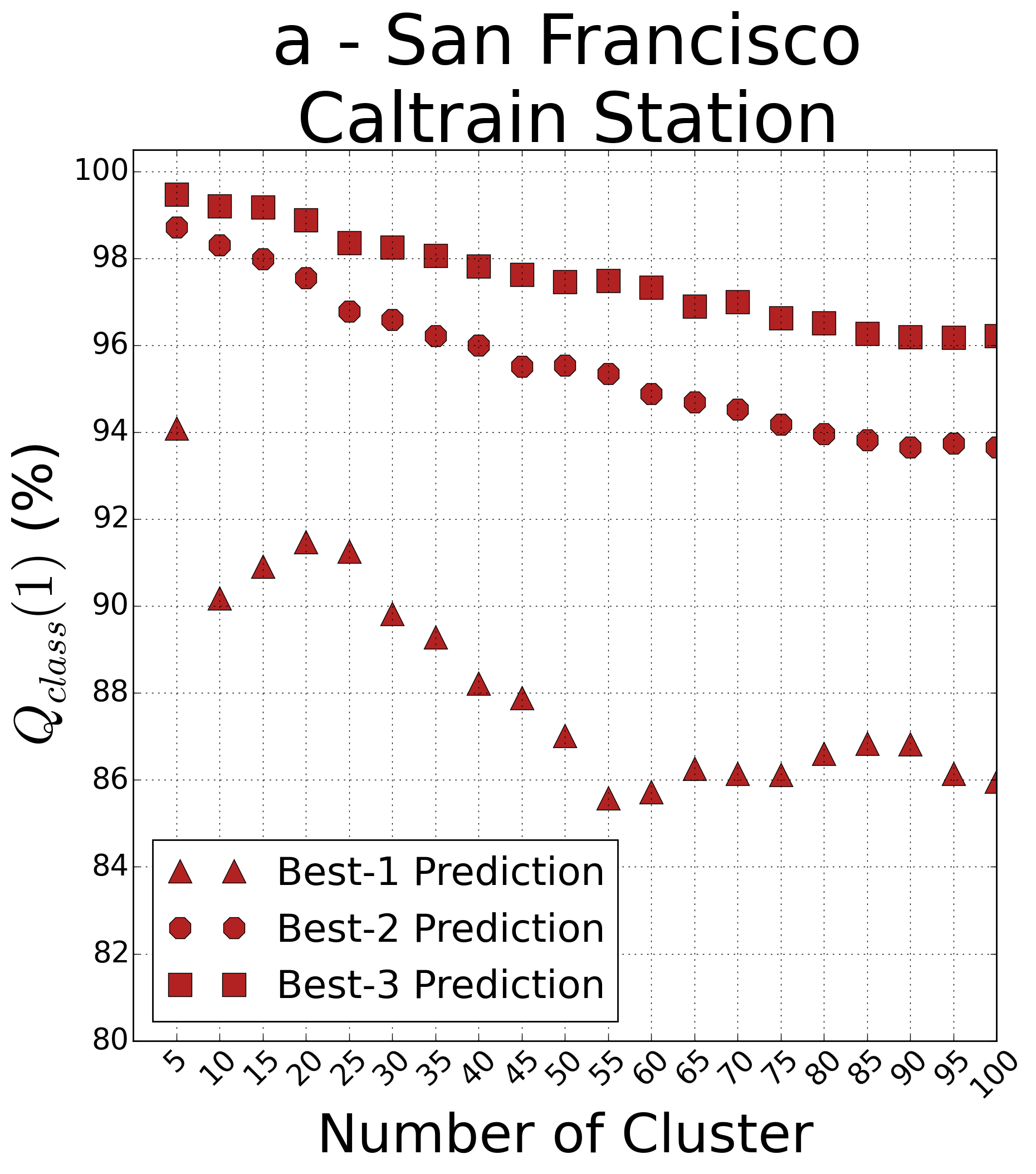}
\end{minipage}
\begin{minipage}{0.49\linewidth}
\includegraphics[width=\textwidth]{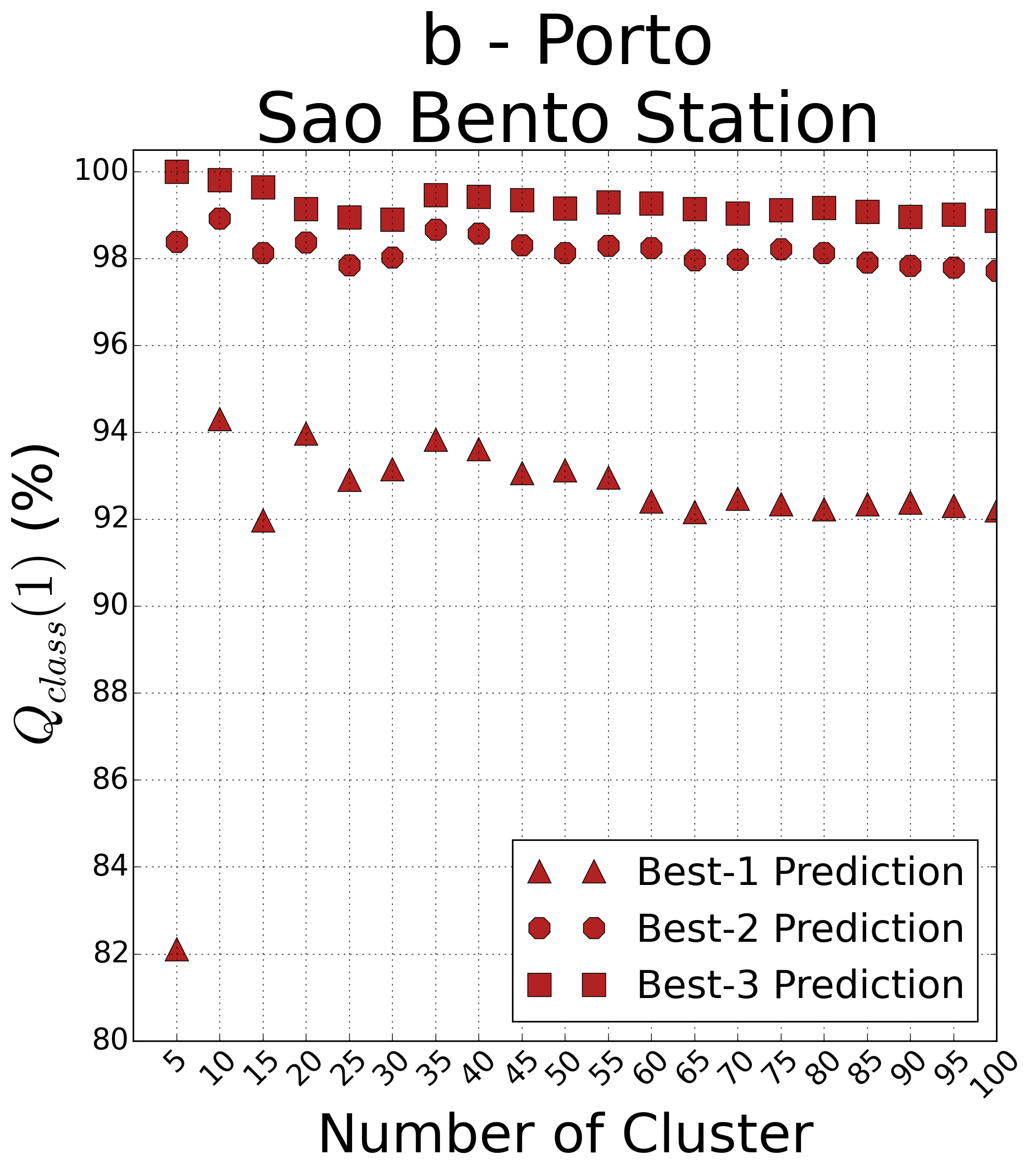}
\end{minipage}
\caption{Percentage of Trajectories Correctly Classified According to Number of Clusters. Compare Best-3 Prediction.}
\label{Classification_Results_best_3}
\end{figure}

In Figure \ref{Classification_Results_best_3}, we can observe the percentage of well classified trajectories, $Q_{class}$ , for $p=1$, \textit{i.e}, when the trips are completed, with respect to the number of clusters of trajectories. Its value does not necessary decrease when the number of clusters increases. When the number of clusters of trajectories is low, the points that compose the trajectory are scattered. Hence the clusters of points found with the Gaussian Mixture have covariance matrices with high value resulting in low likelihoods and low scores. This means that a trajectory can have a low score with respect to its correct cluster. This explains why increasing the number of clusters of trajectories does not always decrease the quality criterion $Q_{class}$. Hence, it can be used as a good criterion to know how well the number of clusters chosen describes the traffic flow of the studied dataset. For the trajectories in San-Francisco, the highest value is $5$ clusters, but the second highest value is obtained for $20$ clusters.  For the trajectories in Porto, the two highest values are found for $20$ and $35$ clusters. 

 We can observe that the percentage of well classified trajectories is always higher than $85\%$ for trajectories in San Francisco and higher than $91\%$ for trajectories in Porto even for $100$ clusters. However, we do not achieve more than $95\%$ of correct classification for both datasets. But our method does not strictly assign one cluster to each trajectory, but a score to each trajectory. We can see that the percentage of correctly classified trajectories significantly increased when we looked at the \textit{best-2} and \textit{best-3} predictions. For trajectories in Porto, this score is almost always greater than $99\%$ for each number of clusters.
 
Hence, It is relevant to look at the score of the correct cluster, and not only the classification rate. For this purpose, we consider the \textit{ROC}(\textit{Receiver operating characteristic}) curves of a one-vs-all classification for every cluster that constitutes a clustering result.

 \begin{figure}
\centering
\begin{minipage}{0.49\linewidth}
\includegraphics[width=\textwidth]{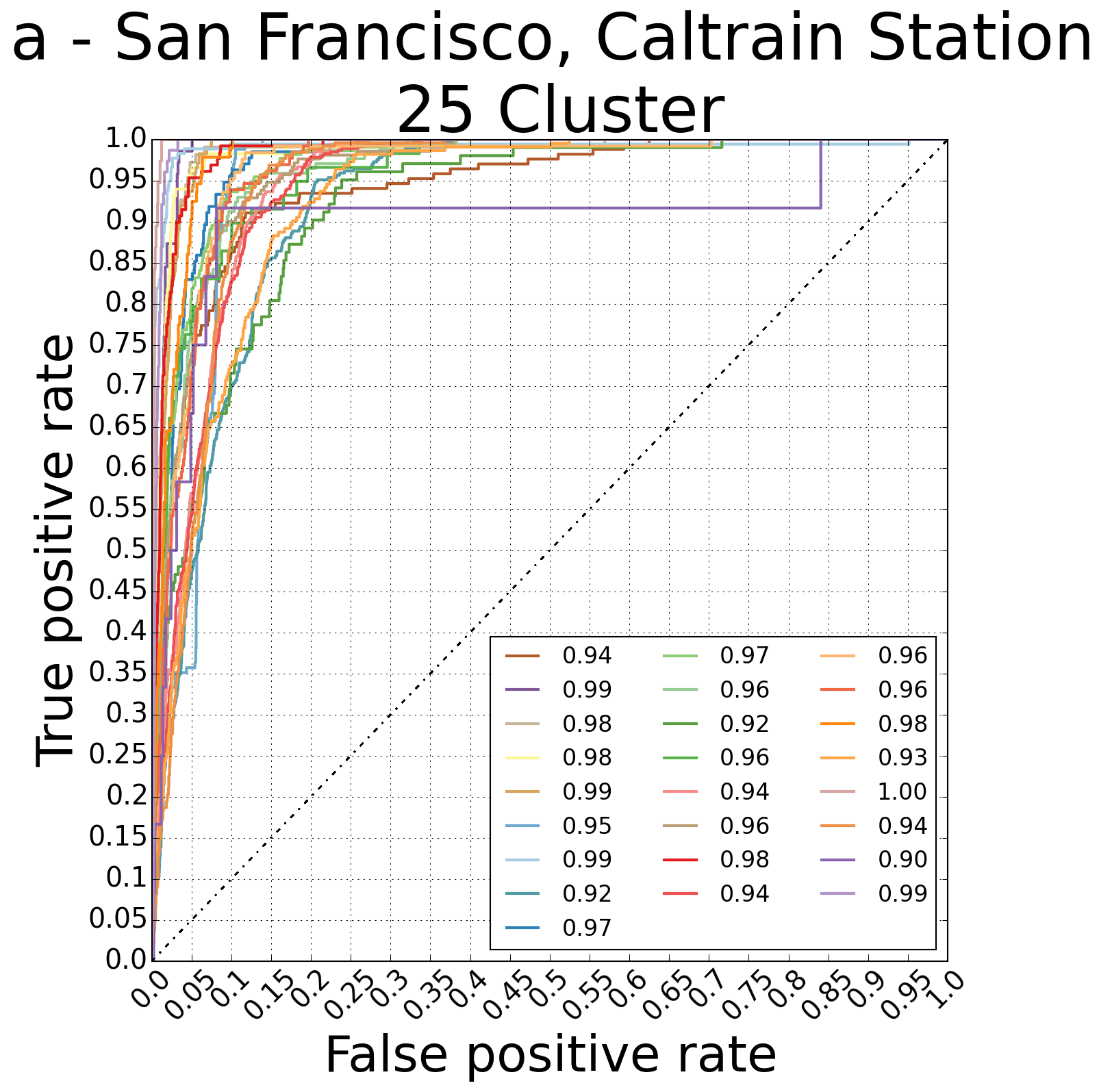}
\end{minipage}
\begin{minipage}{0.49\linewidth}
\includegraphics[width=\textwidth]{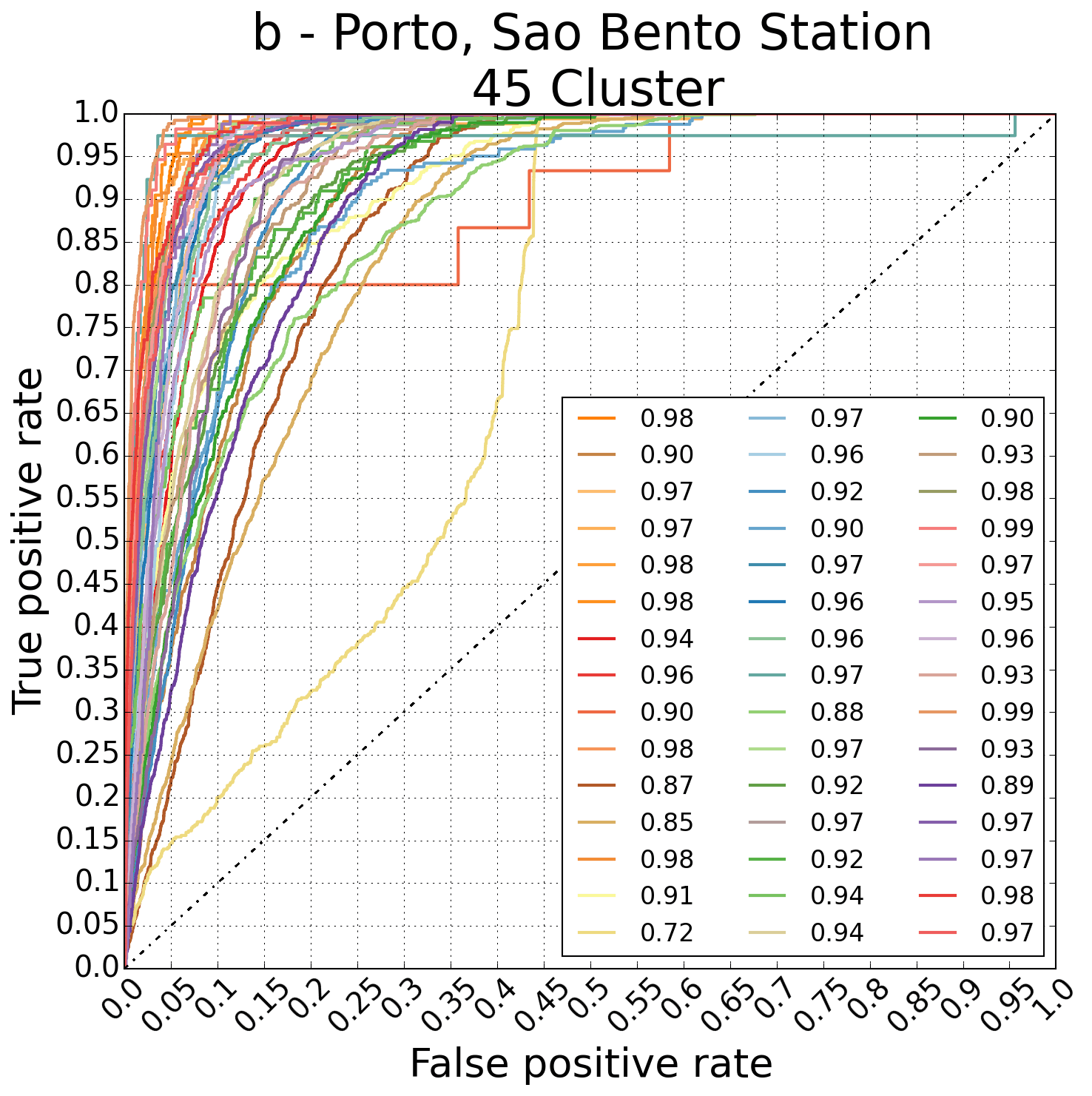}
\end{minipage}
\caption{Roc Curves and AUC for every clusters of trajectories.}
\label{Roc_Results}
\end{figure}

In Figure \ref{Roc_Results}, we can observe these \textit{ROC} curves and their \textit{AUC} (Area Under Curves) for the clustering fixed to $25$ clusters for San Francisco, and $45$ for Porto. All \textit{AUC} are greater than $0.90$, and $17$ are greater than $0.95$ for San Francisco. All but three are greater than $0.90$ and $28$ are greater than $0.95$ for Porto. These results show that even if some trajectories are not assigned to its correct clusters, the \textit{simple scores} for their correct cluster is always elevated.

\subsection{Final Destination Prediction}\label{subsec_fdp}

In this section, we present the results of our method for the prediction of taxi trips destination. To evaluate our method we used the mean of the \textit{Haversine Distance}, which measures distances between two points on Earth based on their latitude and longitude.

\begin{definition}\label{def_haversine} The Haversine Distance, $D_{H}$ between two locations $d1,d2 \in \mathbb{R}^2$ is defined as:

\begin{gather*}
D_{H}(d1,d2) = 2\cdot r\cdot \arctan\bigg(\sqrt{\frac{a}{1-a}}\bigg) \\
a = \sin^2 \bigg(\frac{y_2-y_1}{2}\bigg) + \cos(y_1)\cos(y_2)\sin^2 \bigg(\frac{x_2-x_1}{2}\Big)  \\
\end{gather*}
where $(x_1,y_1)$ and $(x_2,y_2)$ are the longitude and latitude of $d1$ and $d2$ respectively and $R=6371(km)$ is the radius of the Earth. Hence, the Haversine distance returns the distance in Km between two locations on Earth. 
\end{definition}

We can then define the quality criterion for the prediction of the final destination as the mean of the \textit{Haversine} distance between the true location of the final destination, $p^c_{n_c}$, of the trajectory, $T^c$, and the location of the prediction, $d_{pred}(T^c)$.

\begin{definition} The quality criterion, $Q_{pred}$, is defined as :  
 $$Q_{pred}(p) = \frac{ \sum_{T^c \in \mathcal{T}} D_H(d_{pred}(T^c)-p^c_{n_c}) }{\#\mathcal{T}} $$
 \end{definition}
 
 \begin{figure}
\centering
\begin{minipage}{0.49\linewidth}
\includegraphics[width=\textwidth]{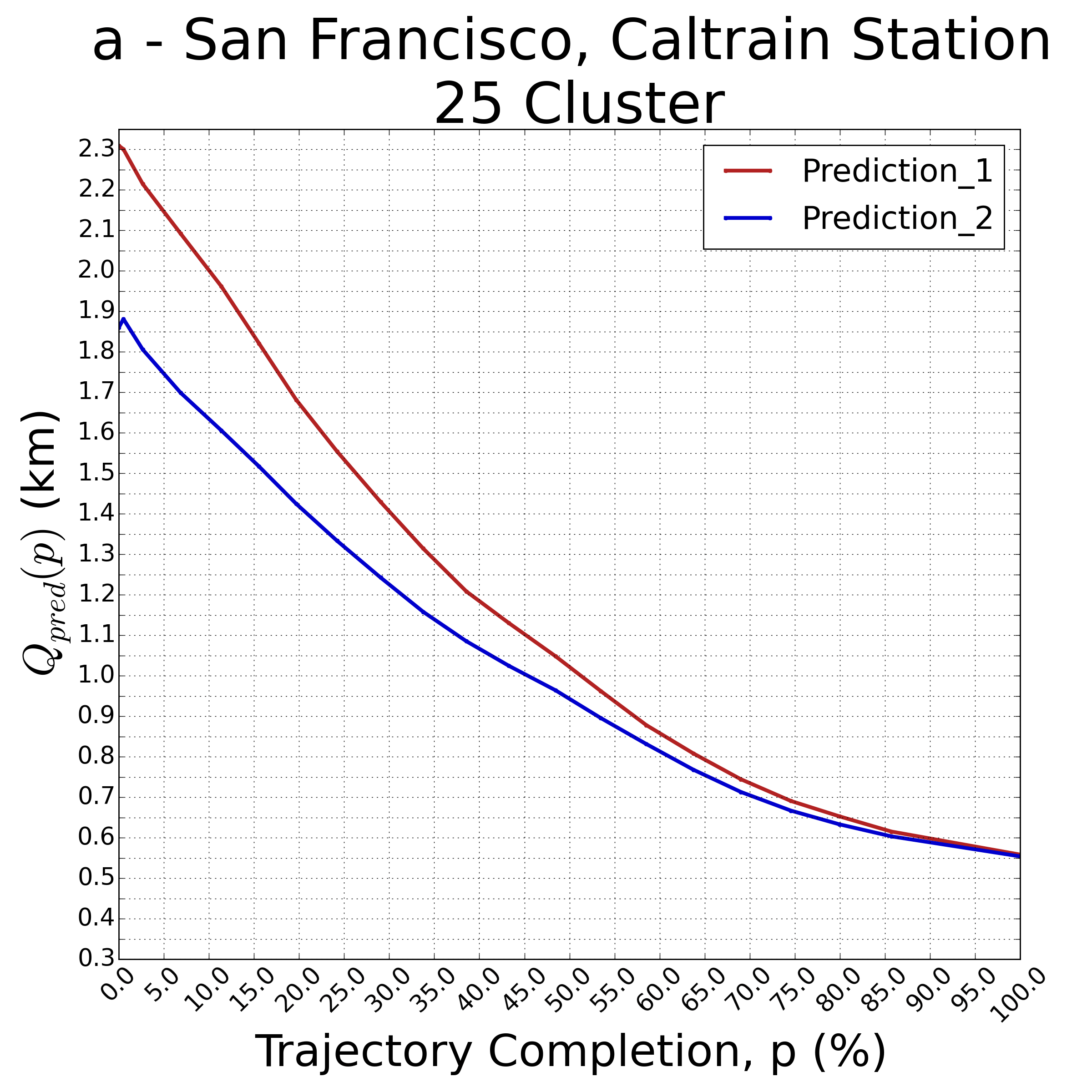}
\end{minipage}
\begin{minipage}{0.49\linewidth}
\includegraphics[width=\textwidth]{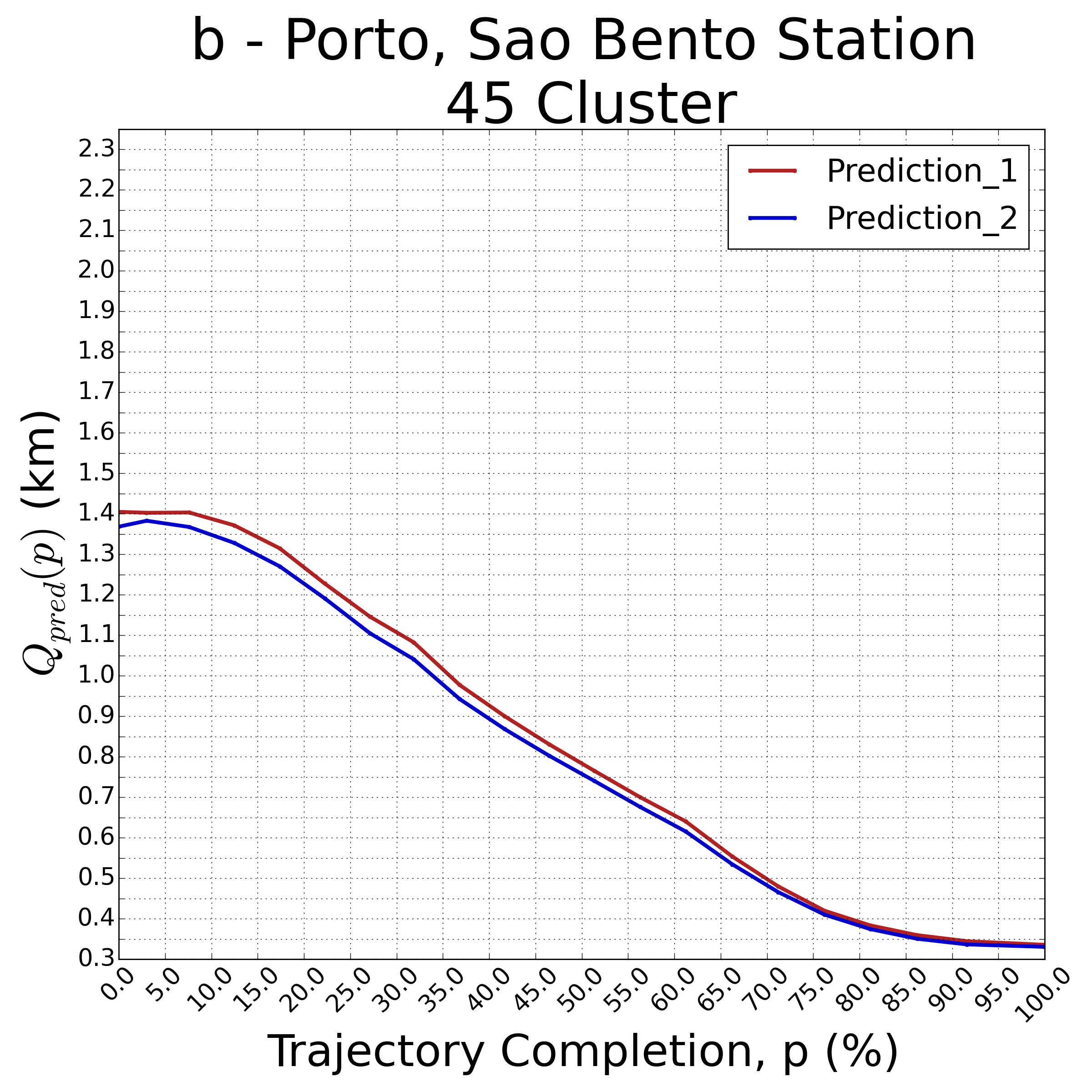}
\end{minipage}
\caption{Mean Error of Final Destination Prediction According to Trajectory Completion. Compare Method 1 and 2.}
\label{fig_destination_prediction_compare_prediction_method}
\end{figure}

\begin{figure}
\centering
\begin{minipage}{0.49\linewidth}
\includegraphics[width=\textwidth]{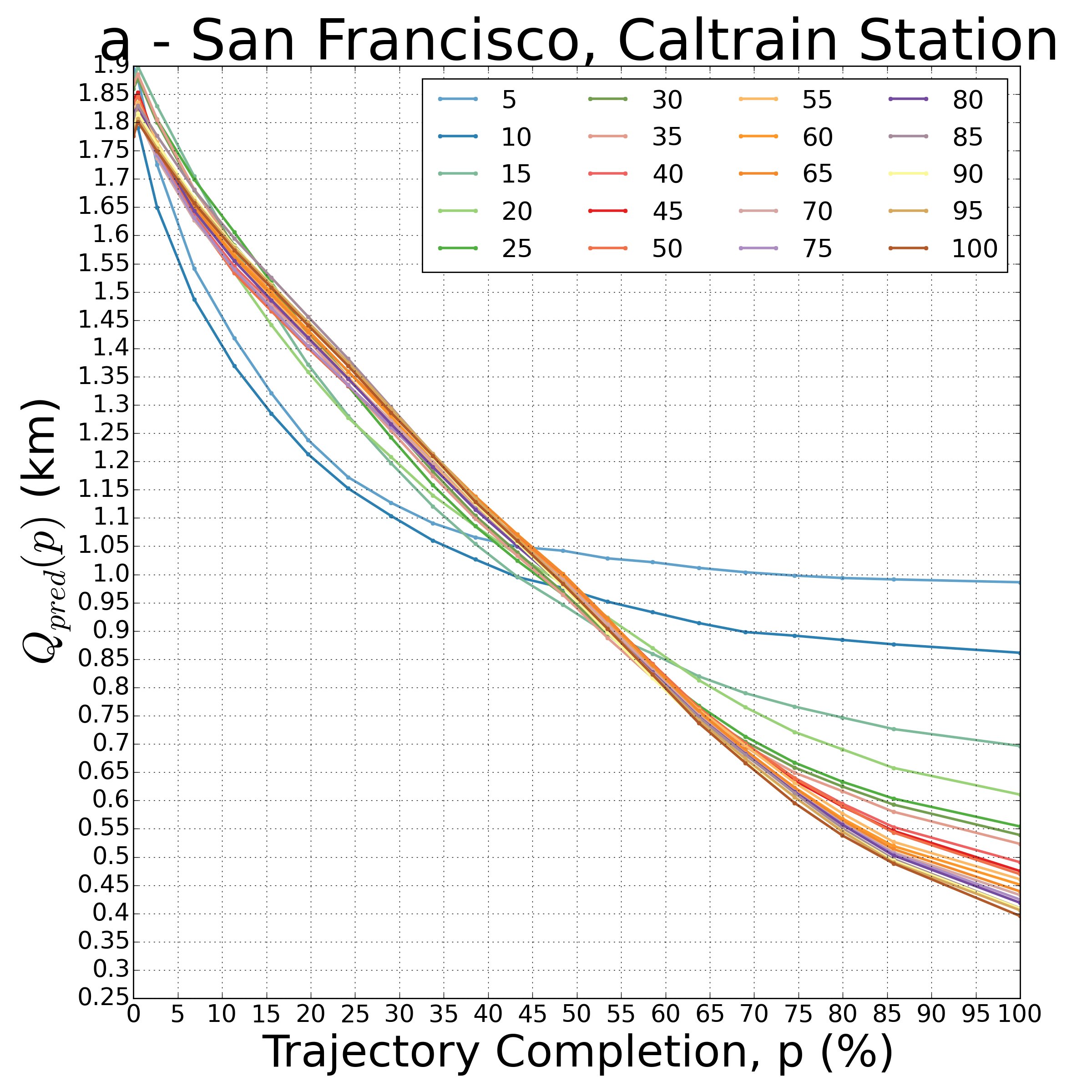}
\end{minipage}
\begin{minipage}{0.49\linewidth}
\includegraphics[width=\textwidth]{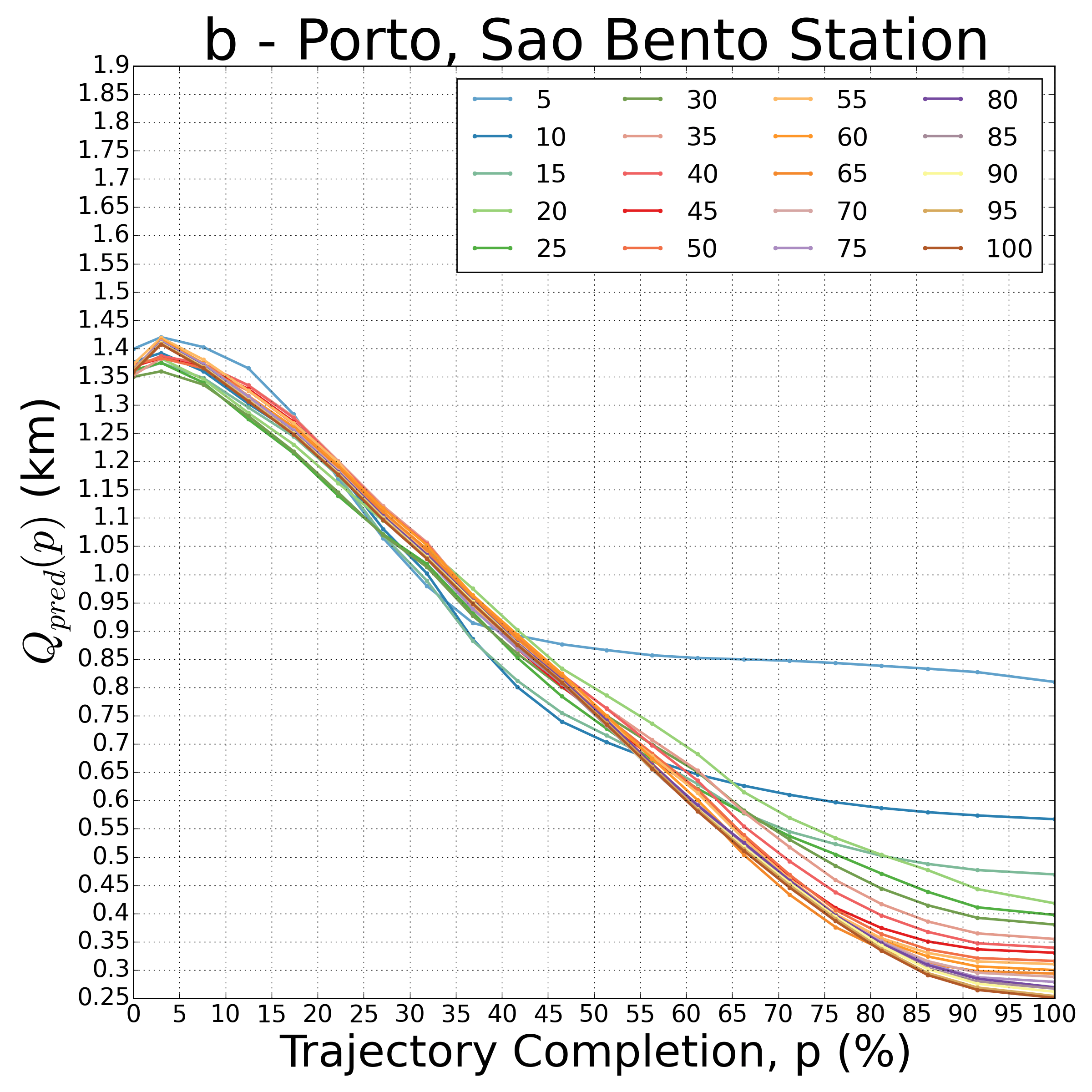}
\end{minipage}
\caption{Mean Error of Final Destination Prediction According to Trajectory Completion. Compare Number of Cluster for Method 2}
\label{fig_destination_prediction_compare_nb_cluster}
\end{figure}

In figure \ref{fig_destination_prediction_compare_prediction_method} we can observe the results of the quality criterion, $Q_{pred}$, according to the trajectory completion. We compare the results of the two prediction methods, $pred_1$ and $pred_2$. Both datasets are displayed, San Francisco(a), on the left and Porto(b), on the right, for a number of clusters of $25$ and $45$ respectively. For San-Francisco, we can observe that the second method gives best results especially at the beginning of the trajectories where $Q_{pred}$ is $400$ meters better using $pred_2$. As the trajectories progress, the results continue to be better with $pred_2$ but the difference between the two methods decreases and after $50\%$ of trajectory completion, the difference is less than $50$ meters. This is expected because the more locations we know for a trajectory, the more confidently we can assign the trajectory to one cluster and one cluster only. Hence, the more locations we know for a trajectory, the more closely the results are using the two methods. For trajectories in Porto, if $Q_{pred}$ also gives better results with $pred_2$, the difference between the method is insignificant. Nevertheless, we will still use the $pred_2$ to compare results according to the number of clusters.

In Figure \ref{fig_destination_prediction_compare_nb_cluster}, we can look at the same quality criterion, $Q_{pred}$,  according to trajectory completion. We display these results for different numbers of clusters from $0$ to $100$. For dataset in San Francsico (a), and for trajectory completion between $0\%$ and $50\%$ the bests results are found for $5$ and $10$ clusters. At these completion rates, the trajectories are more easily assigned to their correct cluster, leading to best results. For completion rates between $50\%$ and $100\%$ the results found with $5$ and $10$ clusters are the worst, because these numbers of clusters do not well enough describe the space. The same conclusion can be made for $15$ and $20$ clusters. UP Until $70\%$ of trajectory completion, there is no strong differences for a number of clusters between $25$ and $100$. When all the trajectories are completed, the more clusters we have, the more precise the prediction is. However, we have a gain of precision of only $200$ meters between $25$ and $100$ clusters. The more clusters of trajectories we have, the more Gaussian Mixture we need to estimate. Hence $25$ clusters of trajectories is the best compromise to well describe the space and to do so within a reasonable computation time. 
For Porto dataset(b), the worst results are found for a number of $5$ clusters, for trajectory completions between $0\%$ and $15\%$ and between $40\%$ and $100\%$. For trajectory completions from $35\%$ to $55\%$, the best results are for a number of clusters of $10$ and $15$, but they yield bad results after $65\%$ and $80\%$ trajectory completions. The same conclusion can be made for a number of clusters between $20$ and $40$. We can observe that the results stabilise when the number of clusters increases from a number of cluster of $45$. The difference of $Q_{pred}$ value for a number of clusters between $45$ and $100$ does not exceed $40$ meters for trajectory completions from $0\%$ to $80\%$. Similarly to San-Francsico, when all the trajectories are completed, the more clusters we have, the more precise the prediction is, but the gain of precision is low, only $100$ meters between  $45$ and $100$ clusters. Hence, $45$ is the best choice for the Porto dataset.

\subsection{Effect On Classification and Prediction}

\begin{figure}
\centering
\begin{minipage}{0.49\linewidth}
\includegraphics[width=\textwidth]{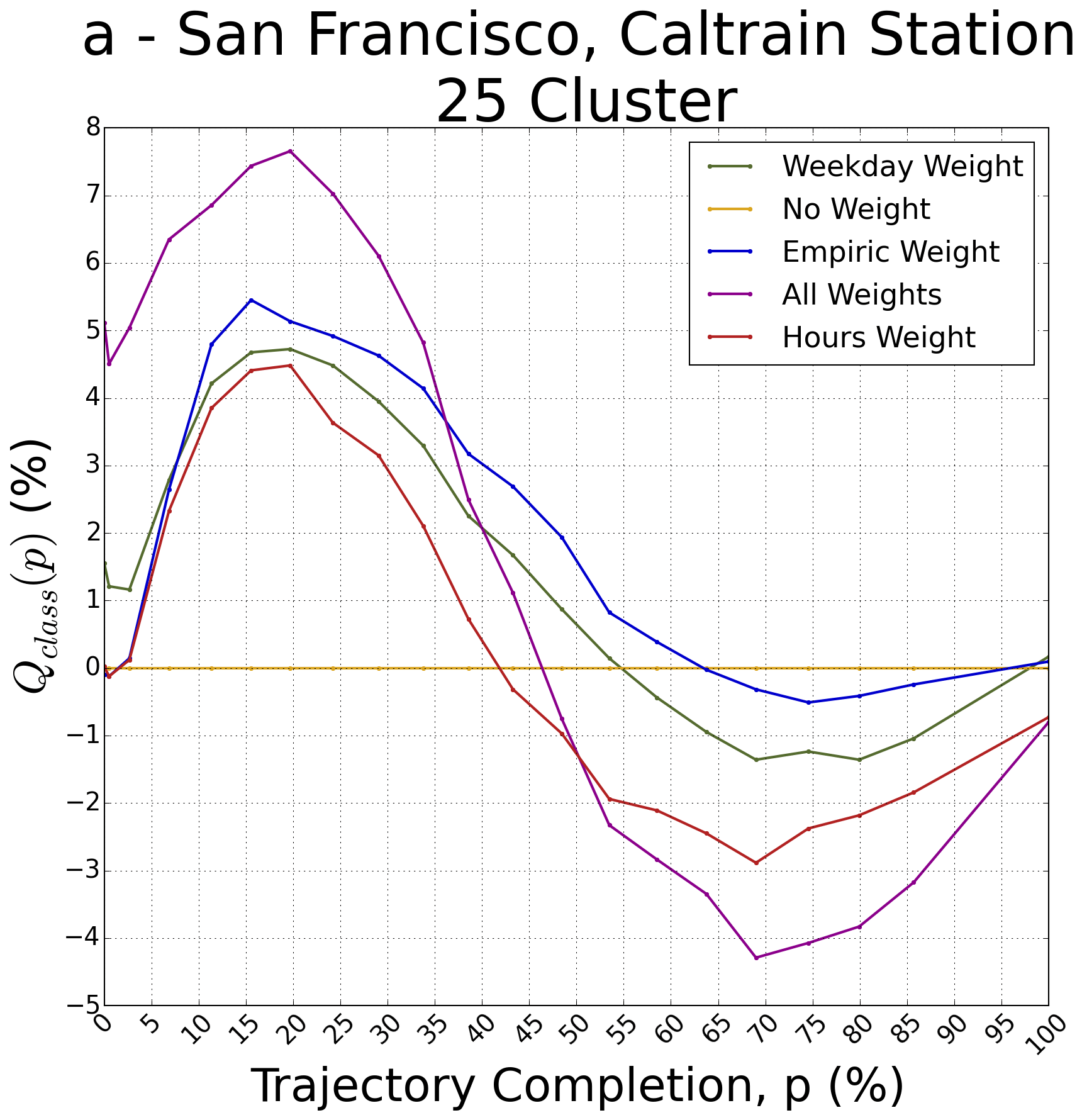}
\end{minipage}
\begin{minipage}{0.49\linewidth}
\includegraphics[width=\textwidth]{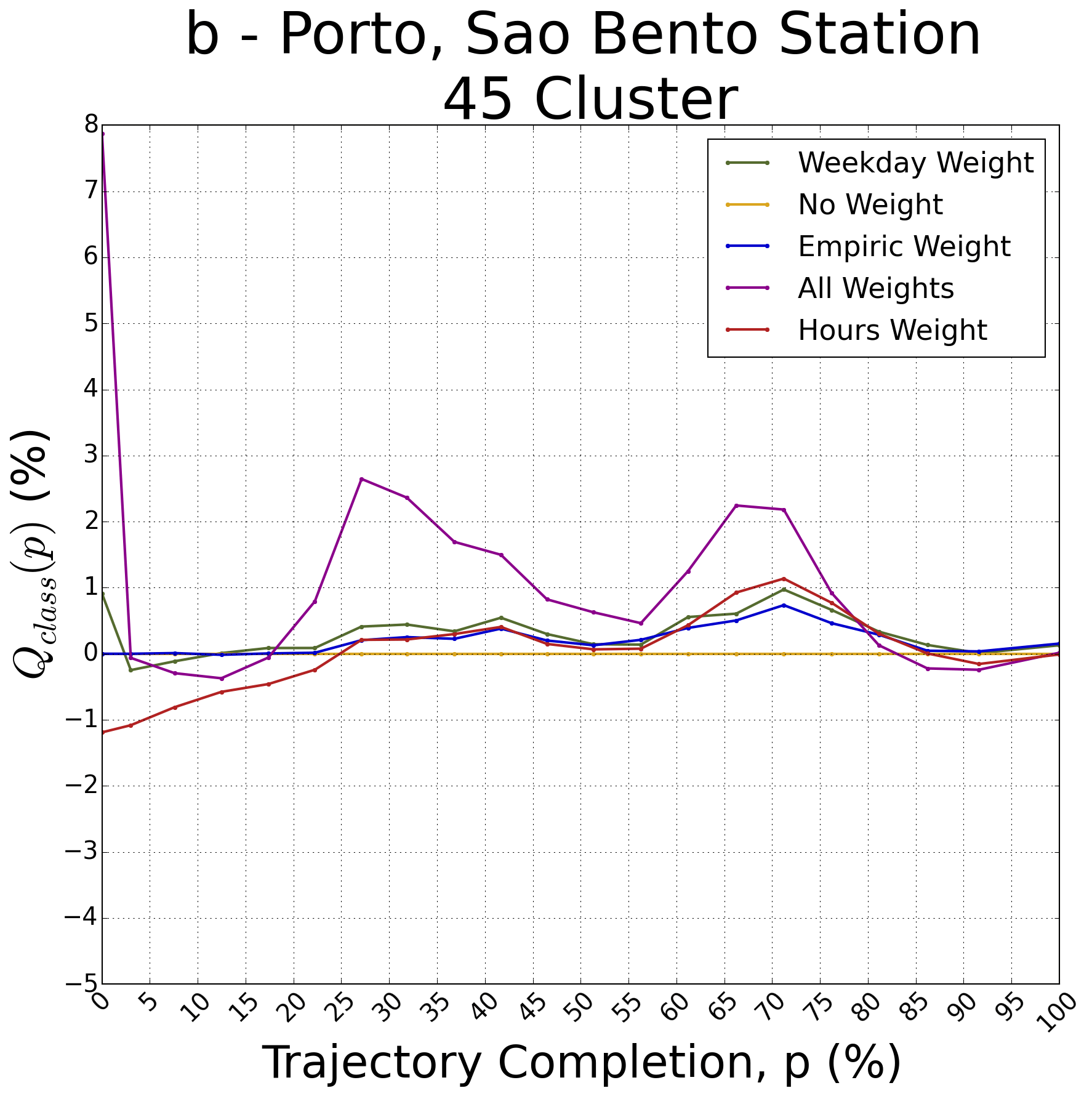}
\end{minipage}
\caption{Improvement of Trajectory Classification With auxiliary Information}
\label{fig_destination_apriori_classification}
\end{figure}

\begin{figure}
\centering
\begin{minipage}{0.49\linewidth}
\includegraphics[width=\textwidth]{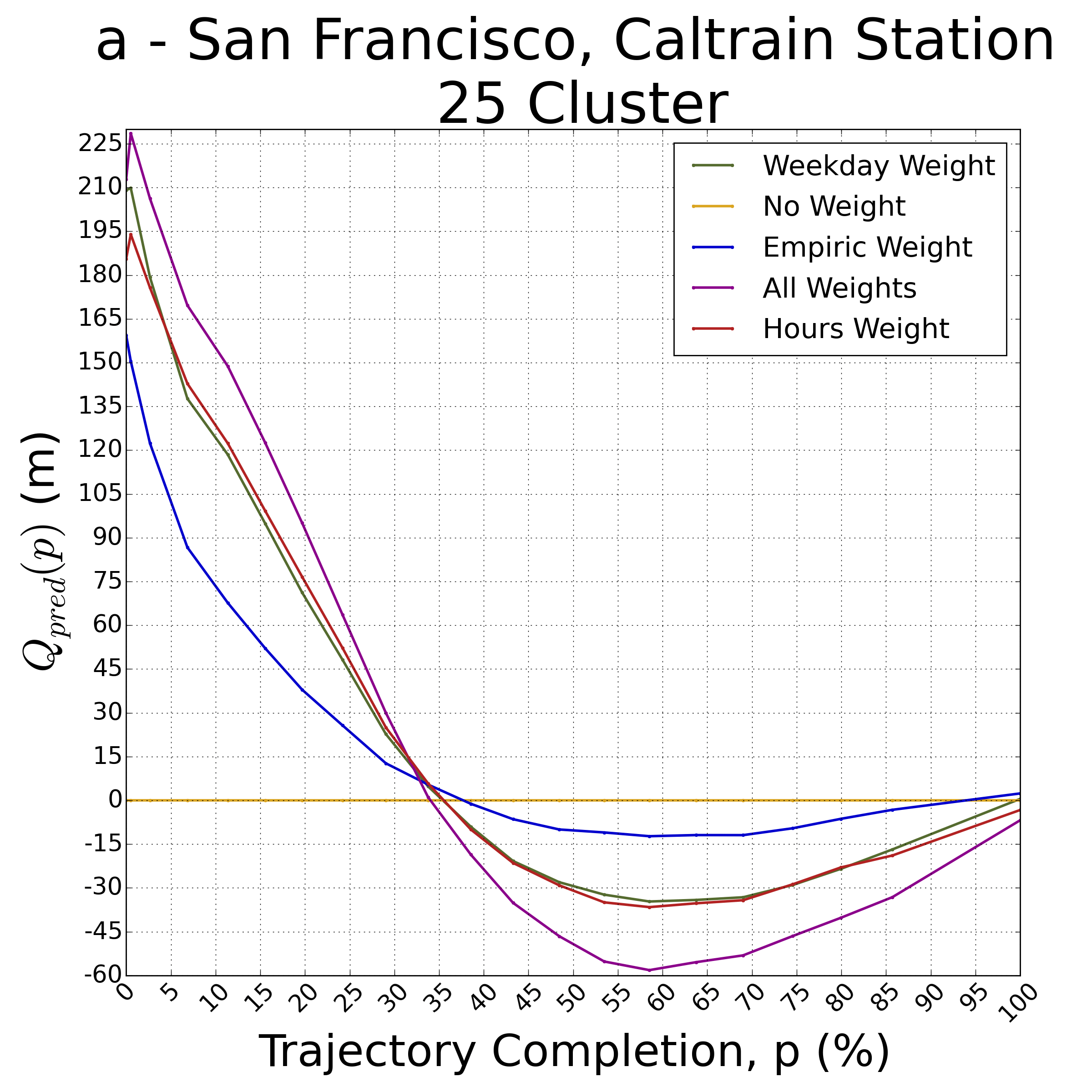}
\end{minipage}
\begin{minipage}{0.49\linewidth}
\includegraphics[width=\textwidth]{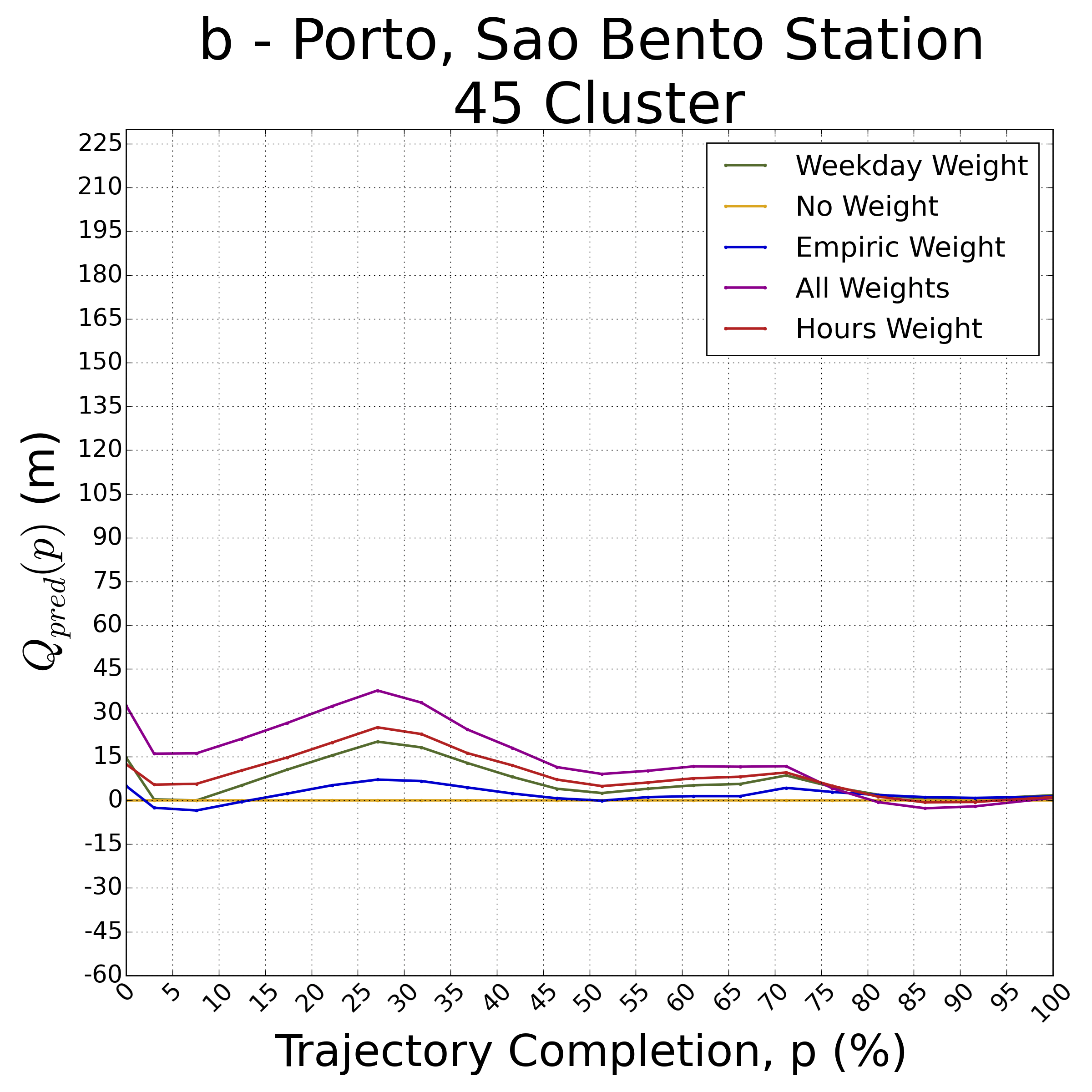}
\end{minipage}
\caption{Improvement of Prediction of Final Destination With auxiliary Information}
\label{fig_destination_apriori_prediction}
\end{figure}

\begin{figure*}
\centering
\includegraphics[width=0.9\textwidth]{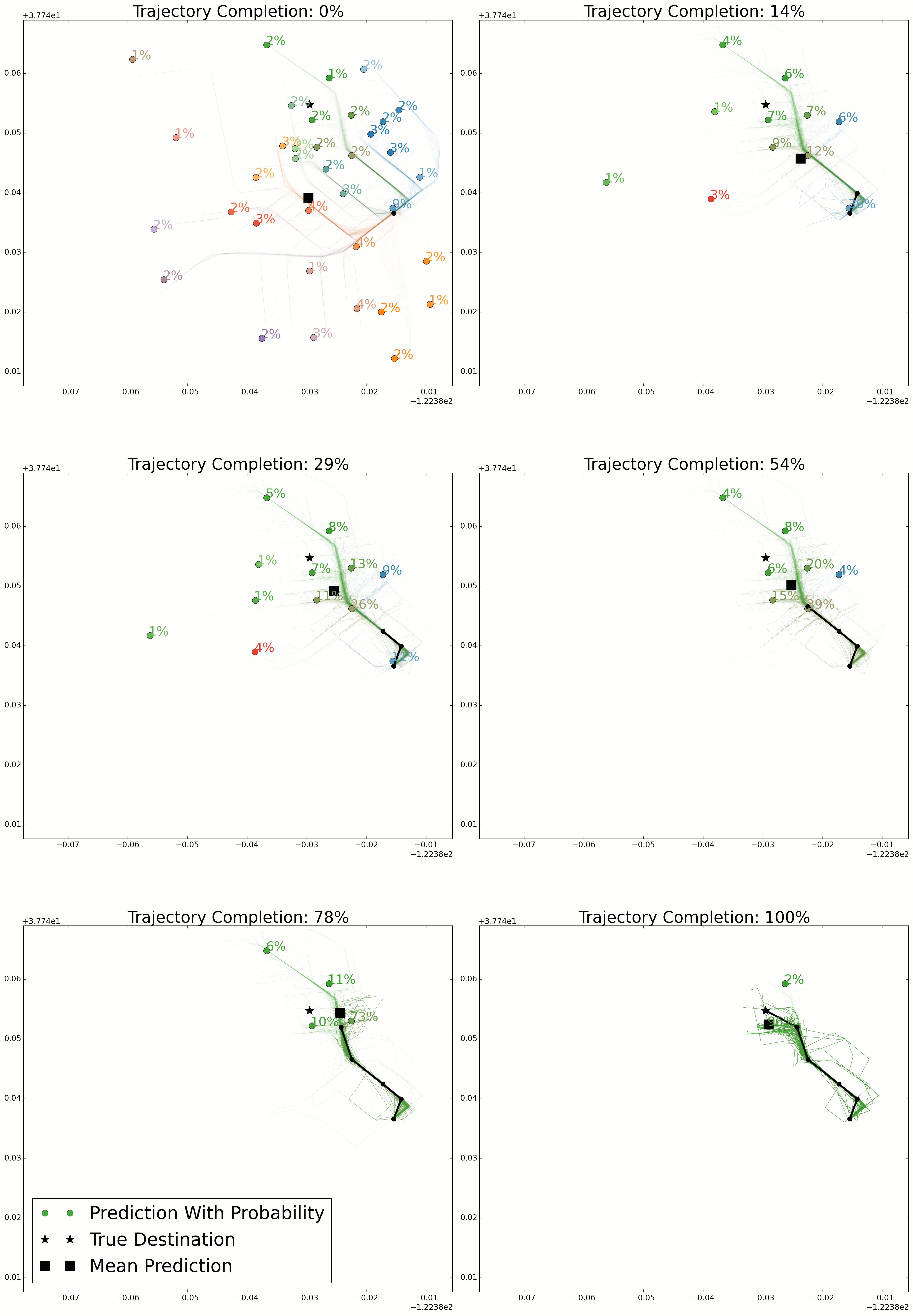}
\caption{Exemple of final destination prediction for a taxi trip In San Francisco}
\label{ffig_pred_example}
\end{figure*}

In Figure \ref{fig_destination_apriori_classification} and Figure  \ref{fig_destination_apriori_prediction}, we observe the effect of different auxiliary weights described in Section \ref{subsection_extended_model} on both the quality criteria $Q_{class}$ and $Q_{pred}$. We display the differences of these criteria with the different weights and the same criteria with no weights according to trajectory completion. For San-Francisco, we display the results for $25$ clusters. We can observe that all weights result in an improvement on both the quality of the classification and the prediction of the final destination in the first part of the trajectories, for trajectory completion between $0\%$ and $35\%-45\%$. The mix of all the weights is yields the best results. The improvement of classification continues until $8\%$ when trajectory completion is at $25\%$ and the improvement of the $Q_{pred}$ criteria is $225$ meters when the trajectory starts and $100$ meters at $25\%$ of trajectory completion. The curves of both quality criteria, $Q_{class}$ and $Q_{pred}$ are not perfectly correlated. This is expected because $Q_{class}$ shows the rate of correct classification, while $Q_{pred}$ displays the prediction quality found with $pred_2$, which uses information of different clusters and not only the first predicted cluster for the prediction. 
Beyond these completion rates, the auxiliary weights deteriorate according to the different quality criteria values. This means that when we have little information about the location of the trajectory, context information help to improve the destination prediction. Whereas when we have sufficient information about the trajectory location, we can confidently predict the correct clusters of trajectories the new trajectory most likely belongs to. Hence, adding auxiliary weight information deteriorates the result. 

The results obtained are different using the Porto dataset. The results for Porto datasets are displayed for $45$ clusters. The different weights improve the prediction, and the mix of all weights yields the best results, but the improvement is always less than $30$ meters which is much less significant than with the San-Francisco dataset. Similarly, the classification is never improved more than $3\%$. Taxi trips in Porto are less influenced by auxiliary variables than taxi trips in San Francisco.

In Conclusion, we have seen that our method gives similar results in trajectory classification, Section \ref{subsec_tc}, and in prediction of final destination, Section \ref{subsec_fdp}, for studied datasets of trajectories in San Francisco and Porto. Taking into account the differences between the structure of the road network of theses two cities proves that our method can be adapted to different datasets, without requiring prior study of the dataset. However, the effect of auxiliary variables is different from one dataset to one another. These results show that the behaviour of the drivers differs from one city to another. It could help traffic managers to better understand the traffic flow of a city. 

We have tested our methods with the test dataset from the Kaggle challenge. This competition has completed but we can still submit an entry to see our score. Our final results produce a mean error of $2,82021$ kilometres, while the best result was $2,03489$ kilometres. Hence we have a difference of $800$ metres relative to the winning solution. This is a promising result because our model has been trained on a sample of trajectories which doest not strictly match the trajectories in the test set. Hence our model can be re-used directly for a different test dataset, and can also be used to predict the destination within the same trajectory, without requiring a new training. In Figure \ref{ffig_pred_example}, we can observe an example of how our model for final destination prediction works. The probability of different possible destination points for a trajectory at 6 different percentages of its trip accomplishment is displayed in this figure. At each moment, the cluster of trajectories with its corresponding final destination and \textit{simple score} are displayed. The more likely the trajectory belongs to a cluster, the more visibly this cluster is displayed on the plot.

\section{Conclusion}\label{Conclusion}
In this paper, we proposed a data-driven method to predict the final destination of vehicle trips using a statistic learning procedure. Vehicle trajectories differ from other trajectories in that they are constrained to a road network, which differs from one place to another, and directly influences the behaviour of the users. The learning step of our method follows a two-step procedure which enables to capture the behaviour of the user. It first models the main paths taken by the users by clustering their complete trajectories. Then, it models main traffic flow patterns within each trajectory's cluster by a mixture of 2d-Gaussian distributions. This yields a data driven grid of locations which describes the all space. 
This model is finally used to predict the final destination of vehicle trips, by assigning the trajectory to the path to whom it belongs the most likely and extracting information from trajectories who follow this path. This prediction is based on the initial location of the trajectory. Since we model the whole path, the prediction can be accomplished at any time during trajectory completion.  
Such method is applied on two different datasets: trajectories of taxi trip moving on two different road networks from San-Francisco, United-State and from Porto, Portugal and proves that such predictions based on the structures of the paths, compete with methods very complex and not easily tractable such as deep learning methods. 
Hence we propose a new description of road traffic that can be used for other research. For example, we can use different information from trajectories inside the clusters to short term prediction, estimate arrival time, or detect abnormal behaviour. Our model provide a better understanding of behaviours of the cars drivers by pointing out the main paths. Hence it can help organise trip distribution of a city.


%
\nocite{*}
\bibliographystyle{IEEEtran}
\bibliography{bibl}

\vspace{-15 mm}

\begin{IEEEbiography}[{\includegraphics[width=1in,height=1.25in,clip,keepaspectratio]{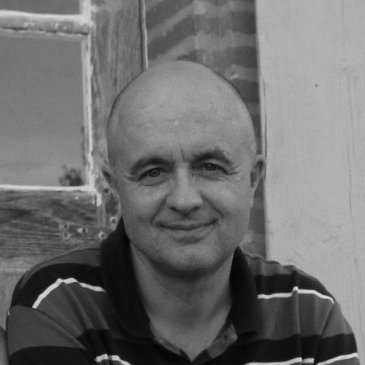}}]{Philippe Besse}
Philippe C. Besse received an Engineering degree in Computer Science from the Polytechnic Institute of Toulouse, France in 1976 and a Ph.D degree in Statistics from the University of Toulouse in 1979.
He is currently a full Professor in the Department of Mathematics of the Institut National des Sciences Appliqu\'ees of Toulouse, having served as Director for the Departement from 2007-2013. Prior to that, he served as the Director of the Laboratory of Statistics and Probabilities, University of Toulouse between 2000 and 2005. He has published more than 50 scientific papers and book chapters in the fields of applied Statistics and Biostatistics.  His research interests include functional data analysis, and the industrial applications of Statistics, Bioinformatics and Data Mining.
\end{IEEEbiography}

\vspace{-12 mm}

\begin{IEEEbiography}[{\includegraphics[width=1in,height=1.25in,clip,keepaspectratio]{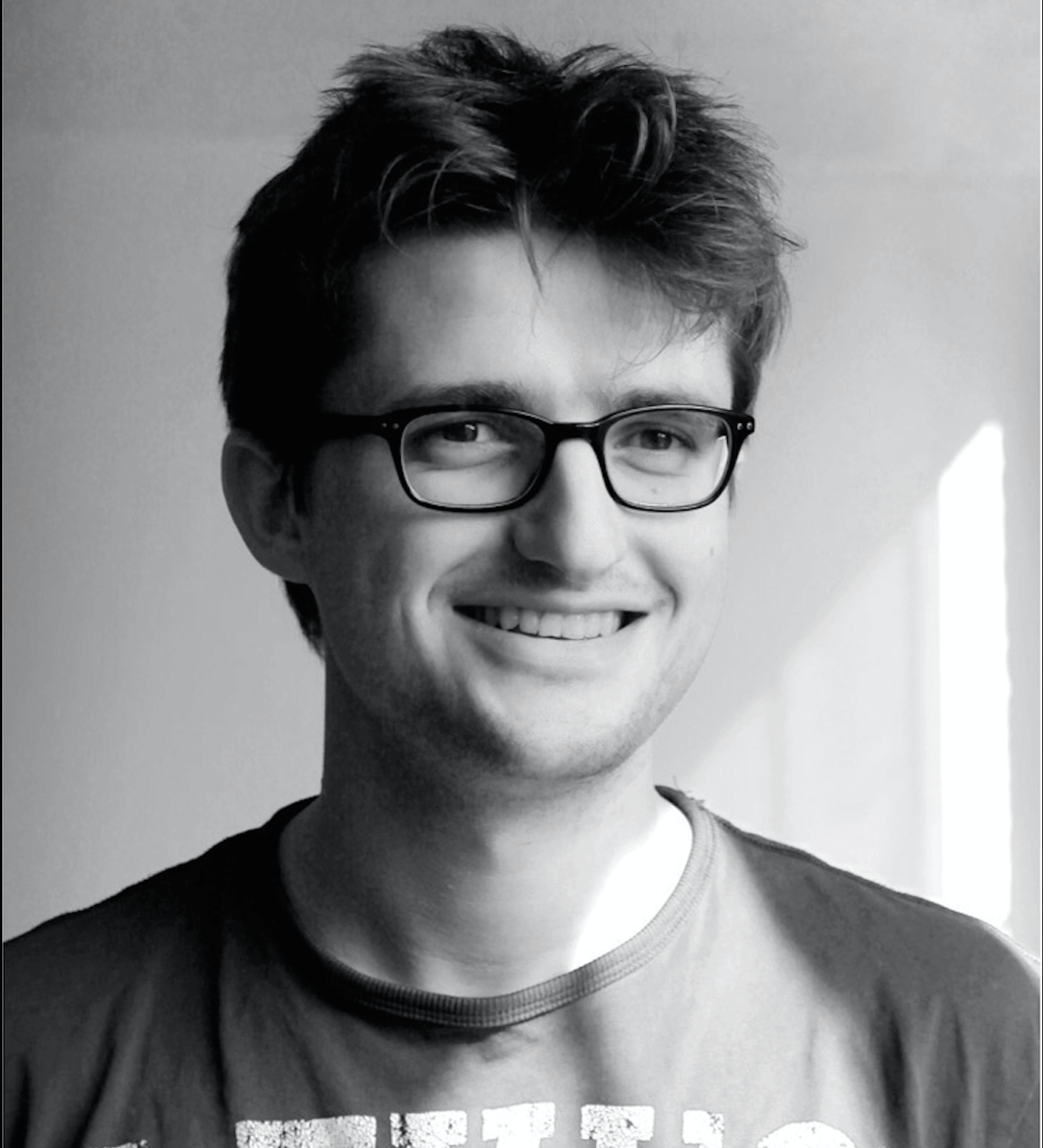}}]{Brendan Guillouet}
Brendan Guillouet received his Engineering degree in Applied Mathematics from the Institut National des Sciences Appliqu\'ees de Toulouse, France in 2013.
He  is  currently  doing a  CIFRE  (Industrial  Training  and  Research) PhD, jointly with Datasio and the Laboratory of Statistics and Probabilities, University of Toulouse. His thesis focuses on Data Mining and Machine Learning methods applied to mobile data.
\end{IEEEbiography}

\vspace{-12 mm}

\begin{IEEEbiography}[{\includegraphics[width=1in,height=1.25in,clip,keepaspectratio]{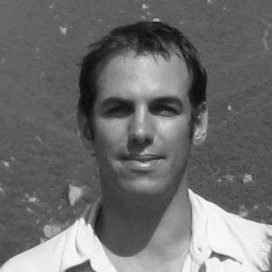}}]{Jean-Michel Loubes}
Jean-Michel Loubes received his Phd of Applied Mathematics at University of Toulouse in 2001. CNRS researcher in statistics at University Paris XI and then Montpellier 2, he is  since 2007 a full Professor in the Institute of Mathematics of the University, having served as Director for the Departement of Statistics and Probability from 2010-2013. He
has published more than 50 scientific papers and book chapters in the fields of applied mathematical Statistics and statistical learning. His research interests include mathematica statistics and  the industrial applications of Statistics, and Machine Learning.
\end{IEEEbiography}

\vspace{-12 mm}

\begin{IEEEbiography}[{\includegraphics[width=1in,height=1.25in,clip,keepaspectratio]{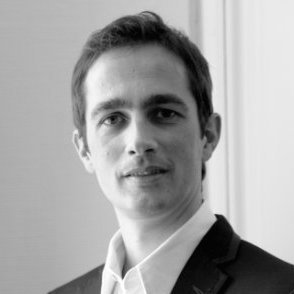}}]{Fran\c{c}ois Royer}
Fran\c{c}ois Royer received his Agronomy Engineering degree from the Ecole Nationale Supérieure Agronomique de Rennes, after pursuing PhD studies in marine ecology at CLS from 2003 to 2005, sponsored by the Centre National des Etudes Spatiales and Ifremer. After a two year post-doctoral position at the Large Pelagics Research Lab at University of New Hampshire, conducting field tagging studies and working on astronomical geolocation algorithms, he moved back to the Oceanography Department of CLS in 2007 where he actively worked on underwater geolocation and Argos positioning.
He is the author of numerous papers specializing in time series analysis, geolocation filtering and smoothing. He founded Datasio in 2012, a private company focusing on Big Data solutions for industrial and environmental applications, where he heads product innovation and commercial development.
His interests range from Bioinformatics and Data Mining to Domain Specific Language development and Functional Programming.
\end{IEEEbiography}

\end{document}